\documentclass{article}
\usepackage{acra}
\usepackage{amsmath}
\usepackage{amssymb}
\usepackage{booktabs}
\usepackage{graphicx}
\usepackage{xcolor}
\usepackage{siunitx}
\usepackage{float}
\usepackage{placeins}

\title{One-Shot Learning for Double-Shot Latte Art: Adapting Cobot Pouring Skills with Dynamic Movement Primitives}

\author{Zain Khan, Victor Hernandez Moreno, and Dinh Tung Le \\
UTS Robotics Institute, School of Electrical, Mechanical and Biomedical Engineering,\\
University of Technology Sydney, Australia\\
\texttt{[Zain.Z.Khan@student.uts.edu.au]}}

\begin{document}
\maketitle

\begin{abstract}
Latte art is a complex task requiring dexterity and precision, making it a challenging and commercially relevant task. A one-shot learning system is employed in which a single barista demonstration is segmented into five functionally distinct phases, each
encoded as an independent Cartesian Dynamic Movement Primitive (CDMP) with its own parameters. Against a single-CDMP baseline, per-phase encoding reduces transition endpoint error by $3.2$--$4.5\times$ and art-stroke shape error by $2.7\times$, reproducing jug tilt to below $1^{\circ}$ RMS in the pouring phases. The learnt pour scales from a
\SI{280}{\milli\litre} to a \SI{400}{\milli\litre} cup through a single spatial offset and a scaled recipe, with no re-demonstration. The phase-based CDMP approach generalises beyond latte art to any
demonstration task whose phases have distinct spatial or temporal requirements whilst demonstrating the ability to adapt reproductions to suit different task requirements.
\end{abstract}

\section{Introduction}
\label{sec:introduction}

\begin{figure}[ht]
    \centering
    \includegraphics[clip, width=1.0\linewidth]{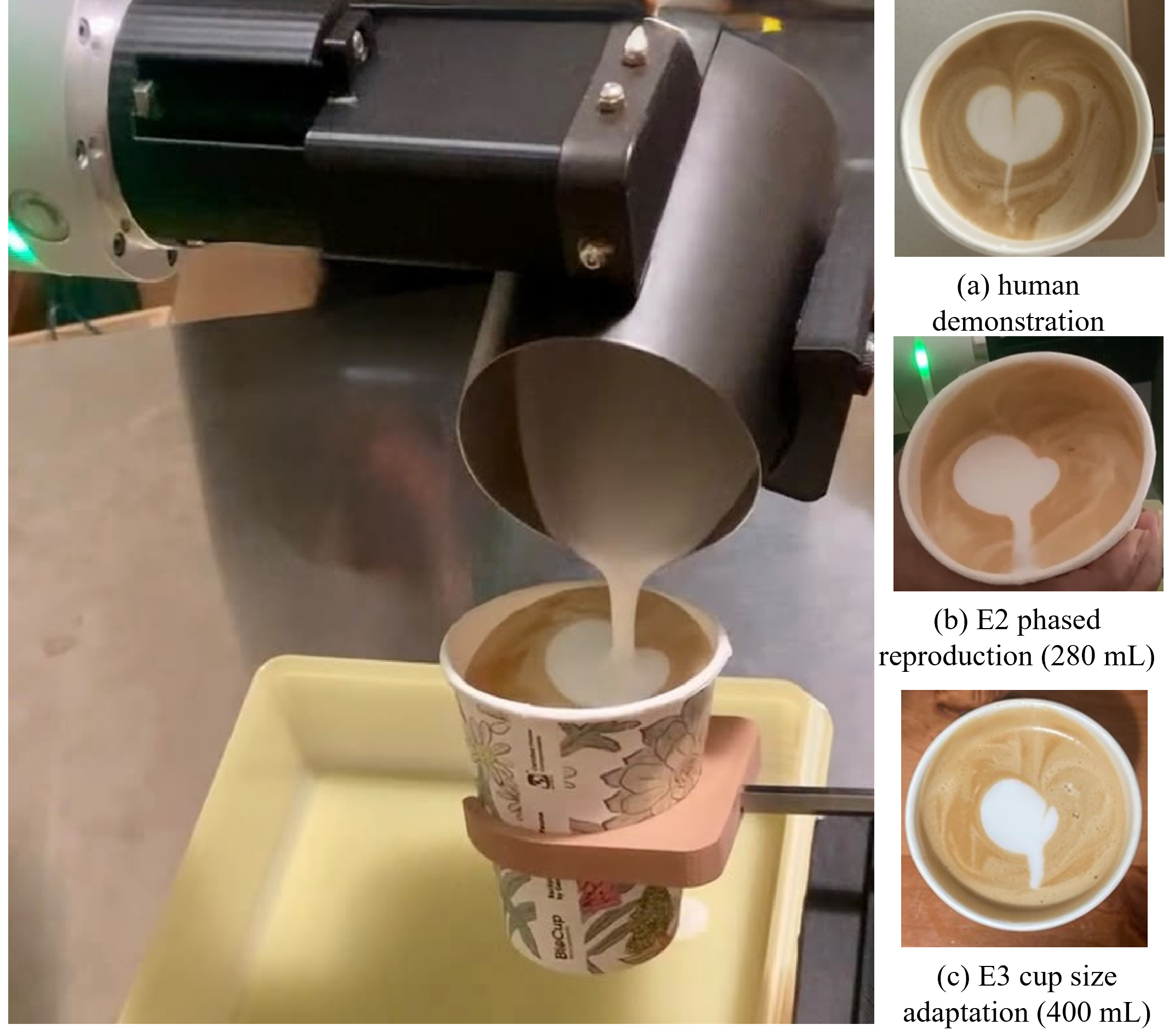}
    \caption{Robot pour in motion with highlight results}
    \label{fig:}
\end{figure}

Coffee preparation is an attractive domain for service robotics because it combines repetitive steps with a product whose quality is dependent on reliable quality. In Australian cafe and coffee shop settings, speed and consistency matter, but social interactions also form a crucial role in contributing to customer experience.
Hospitality research further suggests that personalisation can influence how consumers respond to human--robot collaboration in food preparation~\cite{Kazakova2025Collaboration}. Latte art is a particularly suitable test case: it is recognisable to non-specialists, depends on fine motor coordination, and provides a direct visual stimulus for a customer. 

Latte art is a demanding dexterous task as the appearance of the pattern depends on the relative pose of the jug to the cup, the height of the spout above the liquid surface, jug and cup tilt, motion timing, and the state of the espresso and textured milk. Recent work on complex fluid manipulation also identifies latte art as challenging because it involves viscous textured milk interacting with coffee and because the resulting optimisation landscape is nonlinear and non-smooth~\cite{Xian2023FluidLab}.

Learning from Demonstration (LfD) provides a natural way to acquire such a skill from a human demonstrator when the desired behaviour is easier to demonstrate than to specify analytically~\cite{Siciliano2016Handbook,Ravichandar2020Recent}. Furthermore, the ability to generalise across and beyond demonstration is necessary in service settings, where cups vary in pose, diameter, height, and fill volume.

Dynamic Movement Primitives (DMPs) encode a demonstrated trajectory as a
stable dynamical system with a learnable nonlinear forcing
term~\cite{Ijspeert2013DMP}, and are among the most widely adopted LfD
representations across manipulation domains~\cite{Moreno2024SLR}. Spatial
and temporal parameters can be modified while retaining characteristics of
the demonstrated motion, enabling goal-directed
generalisation~\cite{Pastor2009Learning}. Cartesian DMP formulations
extend this representation to end-effector position and singularity-free
orientation trajectories~\cite{Ude2014Orientation}. These properties make
CDMPs a promising representation for latte art pouring.

This paper investigates how a single human demonstrated latte art pour can be learned using CDMPs and then adapted to different cup sizes, and then reproduced on a collaborative robot without re-demonstration. We make the following contributions:
\begin{itemize}
    \item an autonomous latte-art system that records a human pour with a two-tracker SteamVR setup and transfers the 6-DOF jug trajectory to a Fairino FR5 using CDMP;
    \item a phase-based decomposition of the pour into five functionally distinct phases, each encoded as its own CDMP with independent parameters;
    \item a spatial adaptation that uses the free start and goal of the CDMP to relocate the art phase to a new cup, transferring the demonstration from a \SI{280}{\milli\litre} to a \SI{400}{\milli\litre} cup; and
    \item an end-to-end evaluation directly comparing human and robot pours, and phase-based CDMP against a one-shot CDMP baseline, assessed on real latte art pours with milk and espresso.
\end{itemize}

\section{Related Work}
\label{sec:related_work}

\subsection{Learning from Demonstration and Dynamic Movement Primitives}

LfD encompasses methods that acquire robot policies from examples supplied through kinaesthetic teaching, teleoperation, or passive observation~\cite{Ravichandar2020Recent}. The choice of demonstration interface affects both the naturalness of human motion and robot reproduction quality. Kinaesthetic teaching gives robot compatible joint or end effector data, whereas passive observation can preserve a demonstrator's unconstrained jug handling but requires calibration, pose estimation, and jug mapping.

DMPs represent discrete or rhythmic movements with differential equations whose attractor dynamics are modulated by a learned forcing term~\cite{Ijspeert2013DMP,Saveriano2023Tutorial}. For discrete motion, adapting the start, goal, or temporal scaling can produce a related trajectory without directly editing every demonstration. Early robot experiments showed that start- and goal-parameter can be adapted for manipulation and water serving skills~\cite{Pastor2009Learning}. For Cartesian manipulation, position and orientation require different mathematical treatment because rotations do not form a Euclidean vector space. Ude et al. formulate orientation DMPs using rotation matrices or unit quaternions, retaining DMP modulation and temporal-scaling properties without relying on singular Euler-angle coordinates~\cite{Ude2014Orientation}.

\subsection{Robot Pouring and Generalisation}

Pouring is a well-established benchmark for learning manipulation skills because the task outcome depends on cup geometry, fluid behaviour, and the coupled translation and rotation of the milk jug. Tamosiunaite et al. combined DMP goal learning and shape learning for liquid pouring, demonstrating learning from a human trajectory, real-robot execution, and relearning after a change of bottle~\cite{Tamosiunaite2011Pour}. Compared to existing DMP-based pouring LfD techniques, the novelty of this work comes from the latte art task, cup size generalisation, and its end-to-end evaluation on a real robotic pouring demonstrations.

Prior pouring research also adapts actions to novel object geometry. Brandl et al. infer geometric parameters by warping known point clouds to novel source and receiving containers, then use those parameters to generalise demonstrated pouring actions~\cite{Brandl2014Warped}. Lopez-Guevara et al. use approximate fluid simulation to adapt pouring strategies to sensed novel cup shapes while reducing spillage~\cite{LopezGuevara2017Adaptable}. Other approaches learn predictive pouring dynamics from many demonstrations and refine the skill through self-supervised practice across different cups and fluids~\cite{Huang2021Pouring}. These studies motivate the use of cup geometry as an explicit context variable and provide baselines for discussing what CDMPs do and do not model.

In contrast to water pouring, latte art has a spatially structured outcome produced by interactions between coffee and textured milk. FluidLab introduced simulated latte-art tasks and reported that common model-free reinforcement-learning methods struggled with the high-dimensional fluid state and nonlinear interactions~\cite{Xian2023FluidLab}. 

\section{Method}
\label{sec:method}
 
The pipeline has four stages: an expert barista demonstration is captured as a time series of jug poses and cup-swivel angles; the recording is segmented manually by the barista at demonstration time into functionally distinct phases; each phase is encoded as an independent Cartesian Dynamic Movement Primitive (CDMP); and the learnt primitives are reproduced on a collaborative robot.
 
\subsection{Task Definition and Coordinate Frames}
\label{sec:frames}
 
The task is pouring latte art: steamed milk is poured from a jug into a cup of espresso such that the surface pattern forms a \emph{heart}. Three frames are involved:
 
\paragraph{Task frame $\mathcal{T}$:}
A static \emph{anchor} tracker is rigidly mounted to a demonstration jig with the cup swivel and defines the task frame. Its origin is the anchor tracker; its $+Z$ axis is true vertical, taken from the tracking system's
gravity-aligned room calibration; and its $+X$ axis is the anchor's most-horizontal body axis projected onto the ground plane.
 
Poses are converted from the tracking system's $Y$-up convention to the cobot's $Z$-up convention by a fixed rotation
\begin{equation}
  R_{\mathrm{VR}\rightarrow\mathcal{T}} =
  \begin{bmatrix} 0 & 0 & -1 \\ -1 & 0 & 0 \\ 0 & 1 & 0 \end{bmatrix},
  \qquad \det R_{\mathrm{VR}\rightarrow\mathcal{T}} = +1 ,
\end{equation}
 
\paragraph{Robot base frame $\mathcal{B}$:}
This is the base frame of the cobot and kinematic root of link transformations. The demonstration is recorded in
$\mathcal{T}$ and reproduced in $\mathcal{B}$; the two are related by a rigid transform, calculated as described in Section~\ref{sec:execution}.
 
\paragraph{Tool frame $\mathcal{E}$:}
The jug is mounted to the cobot's tool flange through a 3D printed adapter. During reproduction, a tracker is attached to the jug at the same position as during demonstration, ensuring end-to-end evaluation without transformation offsets. For implementation, a fixed joint is added to the robot's kinematics description, introducing a frame as a child of the final \texttt{wrist3} link:
\begin{equation}
  {}^{\mathrm{wrist3}}\mathbf{p}_{\mathcal{E}} =
  \begin{bmatrix} -0.050 & 0 & 0.240 \end{bmatrix}^{\!\top}\si{\metre}.
\end{equation}
Inverse kinematics is then solved for $\mathcal{E}$.
 
\subsection{Demonstration Acquisition}
\label{sec:acquisition}
 
\paragraph{Jug pose:}
A second tracker is mounted on the jug and sampled at \qty{90}{\hertz}. Each sample is converted to a unit quaternion, transformed into $\mathcal{T}$, and expressed relative to the anchor.
 
\paragraph{Cup swivel angle:}
The cup swivel angle is recorded alongside the jug pose, as the barista
rotates the cup throughout the pour and its angle contributes to how the
pattern forms.
 
\paragraph{Phase marking:}
The barista marks phase transitions during the pour using a foot pedal,
leaving both hands free for the jug and cup swivel. Phase transitions are
predetermined and the barista learns them before recording, so phases are
labelled during demonstration rather than segmented after the demonstration. The
deliberate act of pressing the pedal also reinforces near-zero jug motion
at each transition, which the CDMP sequencing relies on as a condition for
start and goal poses.
 
\paragraph{Preprocessing:}
Recordings are processed offline before fitting. Idle head and tail regions are trimmed and outliers are rejected independently per sensor: jug pose samples exceeding per-frame position or rotation jump limits are replaced by interpolation and SLERP respectively, while the cup swivel channel is averaged against a rolling median. Position and cup angle are smoothed with a Savitzky--Golay filter; orientation is smoothed in the quaternion log space about the first sample, keeping the tangent-space approximation valid. Finally all channels are resampled onto a uniform \qty{50}{\hertz} grid, positions by cubic spline, orientations by SLERP, and phase labels by nearest neighbour.
 
\subsection{Phase Decomposition}
\label{sec:phases}
 
A complete pour comprises of segments whose spatial characteristics, kinematics and functional roles differ. Therefore, each demonstration is segmented into five phases, summarised in Table~\ref{tab:phases}.
 
\begin{table}[h]
\centering
\small
\caption{Phase decomposition of a latte art demonstration}
\label{tab:phases}
\begin{tabular}{llr}
\toprule
\# & Phase & Type \\
\midrule
1 & \texttt{base\_to\_mix} & transition \\
2 & \texttt{mixing}        & pouring    \\
3 & \texttt{mix\_to\_art}  & transition \\
4 & \texttt{art}           & pouring    \\
5 & \texttt{return\_base}  & transition \\
\bottomrule
\end{tabular}
\end{table}
 
\begin{figure*}[ht]
\centering
\includegraphics[width=0.9\textwidth]{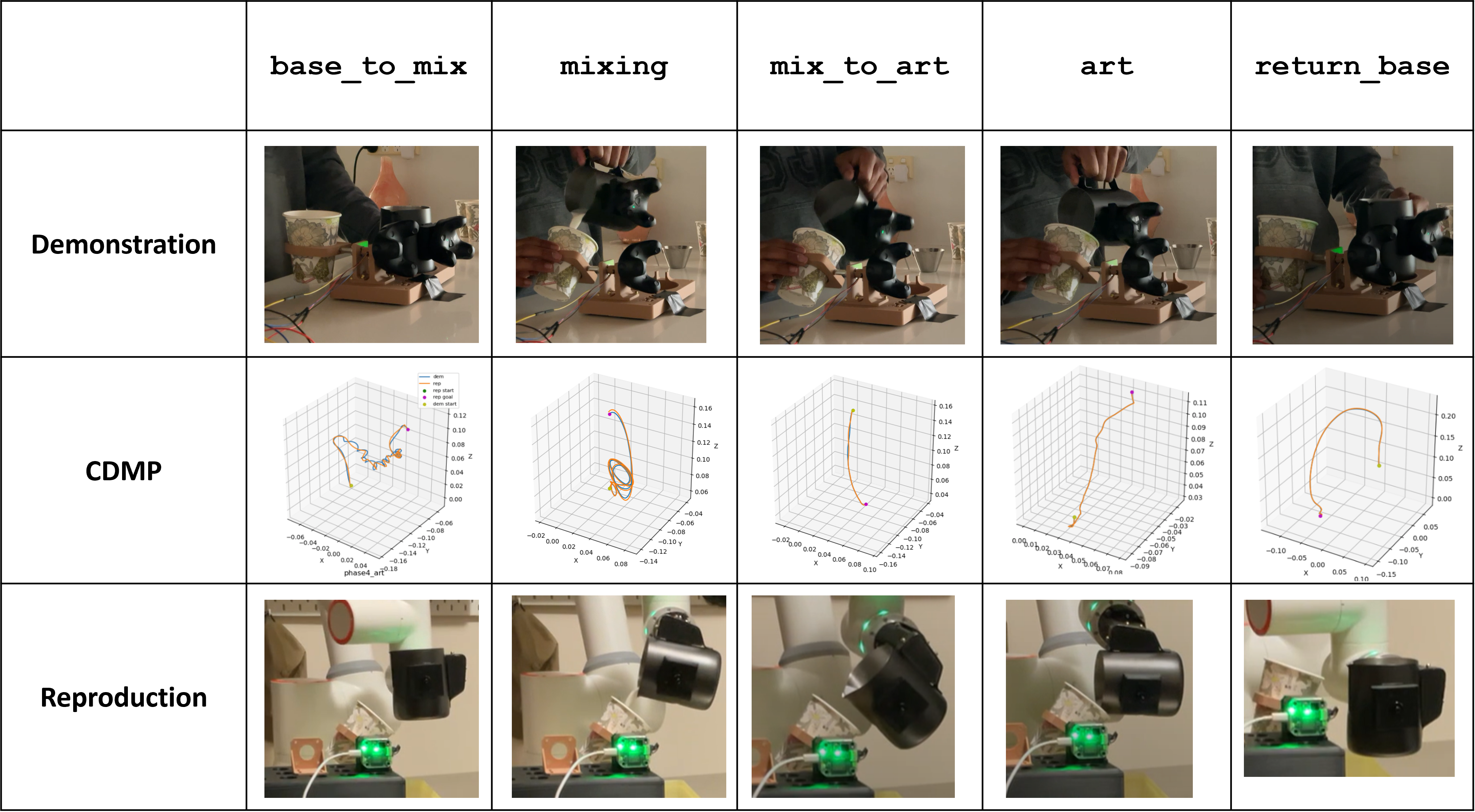}
\caption{Phase decomposition shown across each layer.
Each column is one phase; rows show the human demonstration, the learnt primitives, and the robot reproduction.}
\label{fig:phase-grid}
\end{figure*}

Phase 1~(\texttt{base\_to\_mix}) and 5~(\texttt{return\_base}) are start and end transitions respectively: the jug is repositioned and no milk is poured, so only the endpoint is relevant. Phases 2~(\texttt{mixing}) sets the canvas (the base mixture of espresso and milk that latte art flows onto) by circling to distribute milk beneath the crema; its value lies in the area covered whilst adding volume to set the canvas for the proceeding the art phase.
Phase 3~(\texttt{mix\_to\_art}) is a crucial transition motion between the \texttt{mixing} and \texttt{art} phases where no pouring occurs and the jug is repositioned to produce art on the surface.  
Phase 4~(\texttt{art}) forms the pattern: the jug pours milk near the surface of the crema to build the round body of the heart and then finishes with a single pull-through stroke. See Figure~\ref{fig:phase-grid} for a visual illustration of the phases.
 
\subsection{Cartesian Dynamic Movement Primitives}
\label{sec:cdmp}
 
Each phase is encoded independently as a CDMP comprising transformation systems for position and orientation and a shared canonical system.
 
A phase variable $s$ decays monotonically from $1$ to $0$, decoupling the
primitive from wall-clock time,
\begin{equation}
  \tau \dot{s} = -\alpha_s s ,
\end{equation}
with the temporal scaling factor~$\tau$ and the decay rate~$\alpha_s$.
With the position~$\mathbf{p}$, the start~$\mathbf{p}_0$, the goal position~$\mathbf{p}_g$, the stiffness~$K$ and critical damping~$D = 2\sqrt{K}$, the position transformation system is
\begin{equation}
  \tau^2 \ddot{\mathbf{p}} = K(\mathbf{p}_g-\mathbf{p})
    - D \tau \dot{\mathbf{p}}
    - K(\mathbf{p}_g-\mathbf{p}_0)s
    + K \mathbf{f}_p(s) .
  \label{eq:pos-dmp}
\end{equation}
Orientation is represented by a unit quaternion~$\mathbf{q}$ and encoded with the formulation of Ude et al.~\cite{Ude2014Orientation}, in which the error to the goal orientation~$\mathbf{q}_g$ is expressed in the tangent space through the quaternion logarithm,
\begin{equation}
  \mathbf{e}_q = 2\log(\mathbf{q}_g \otimes \bar{\mathbf{q}}),
\end{equation}
and evolves under a transformation system of the same form as (\ref{eq:pos-dmp}). Integrating in the tangent space and mapping back through the exponential keeps the reproduced orientation on the unit sphere without renormalisation. The pose goal of a phase is therefore the pair $(\mathbf{p}_g, \mathbf{q}_g)$.
 
Both transformation systems are driven by normalised radial basis function (RBF) expansions gated by the phase variable,
\begin{equation}
  \mathbf{f}(s) = \frac{\sum_{i=1}^{N} \psi_i(s)\, \mathbf{w}_i}
                       {\sum_{i=1}^{N} \psi_i(s)} \, s ,
  \qquad
  \psi_i(s) = e^{-h_i (s - c_i)^2},
\end{equation}
whose weights $\mathbf{w}_i$ are learned from the demonstration by least squares. For convenience we set the number of basis functions per phase through a density parameter~$\rho$ (basis functions per second), so that
$N = \max(10, \lceil \rho \, \tau_{\mathrm{dem}} \rceil)$ for a phase of demonstrated duration~$\tau_{\mathrm{dem}}$. This makes the resolution of a phase an explicit parameter rather than a consequence of its length: a single primitive fitted over a whole pour distributes basis functions uniformly in time, and so under-resolves precisely those phases that contain the most spatial detail per unit time.

\paragraph{Sequencing:}
For merging sequential DMPs, the first approach was applied ~\cite{Saveriano2019Merging}.
Each phase is integrated to convergence and the next initialised at the
terminal state of its predecessor. The phases
are contiguous segments of a single recording, so each phase's pose goal
coincides with the following phase's start, and no change to the DMP
structure is required. A residual gap remains at each boundary because the
forcing term vanishes only asymptotically; we measure this gap and
translate each succeeding phase to close it, and reject a reproduction if
any gap exceeds a set threshold.

\paragraph{Temporal scaling:}
Scaling $\tau$ changes the execution time without altering the
demonstrated path. For pouring phases this also changes the poured
volume. Only transition phases are therefore temporally scaled.

\subsection{Reproduction and Execution}
\label{sec:execution}

\paragraph{Frame anchoring:}
The trajectory is anchored to the robot's current pose at execution time.
Position is translated so the first commanded pose coincides with the
current end-effector pose. Orientation is aligned by rotation about the
vertical axis only, obtained by swing--twist decomposition, since both
frames are gravity-aligned and the true frame offset is a pure yaw.

\paragraph{Tool orientation:}
The body-frame correction between the tracker and the jug axis is
recovered from the jug being upright at the first demonstration sample and
upright on the flange at execution time. This fixes two rotational degrees
of freedom; the remaining spin about the jug axis determines spout heading
and is set once per mounting.

\paragraph{Phase chaining:}
One shared transform is applied to all phases, preserving inter-phase
geometry. Residual gaps at phase boundaries are measured and closed; a
reproduction is aborted if any gap exceeds a set threshold.

\paragraph{Joint-space execution:}
Cartesian waypoints are converted to joint angles by per-waypoint inverse
kinematics, each solve seeded with the previous solution to maintain a
single IK branch. The trajectory is sent as a single timed joint
trajectory, validated against joint limits before execution begins.

\paragraph{Cup reproduction:}
The recorded cup angle is resampled per phase onto the executed
trajectory's timeline so that phases with differing $\tau$ remain
synchronised, then concatenated and resampled onto a uniform grid. The
complete profile is uploaded to the motor controller before motion begins
and released by a single trigger.
 
\subsection{Spatial Generalisation Across Cup Sizes}
\label{sec:generalisation}
 
A demonstration recorded for one cup size can be transferred to another, being able to adapt a pour to multiple cup sizes without re-demonstration. The CDMP formulation admits this directly, because the start and goal of each phase are free parameters: the learnt forcing term shapes the trajectory between them, and displacing them displaces the reproduction whilst remaining topologically similar.
 
For cup adaptation the relevant control is the goal of the transition phase preceding the art phase. Because subsequent phases are chained to their predecessor's terminal pose, displacing that goal relocates the art
phase and everything after it as a unit.
 
Let $d_{\mathrm{ref}}$ be the rim diameter of the reference cup, $d$ that of the target, $\mathbf{c}$ the cup centre in $\mathcal{T}$, and $\bar{\mathbf{p}}_{\mathrm{art}}$ the centroid of the demonstrated art phase. We express the required offset as a recentring term plus a radial scaling term,
\begin{equation}
  \Delta\mathbf{g} =
  \underbrace{\left(\mathbf{c} - \mathbf{c}_{\mathrm{ref}}\right)}
             _{\text{recentring}}
  + \underbrace{\left(\frac{d}{d_{\mathrm{ref}}} - 1\right)
    \left(\bar{\mathbf{p}}_{\mathrm{art}} -
          \mathbf{c}_{\mathrm{ref}}\right)}_{\text{radial scaling}} .
  \label{eq:cup-offset}
\end{equation}
The first term places the pattern at the centre of the new cup; the second scales the pattern's radial extent so that it occupies the same fraction of the rim. In the experiments reported here the cups were mounted concentrically and the pattern was recentred without radial rescaling, so only the first term was applied. The recentring offset was found by iterative adjustment, guided by the phase-footprint diagnostic reported by the reproduction pipeline, converging on
$\Delta\mathbf{g} = (0, +7, 0)\,\si{\milli\metre}$ in $\mathcal{T}$.
The correction is purely lateral, consistent with a horizontal recentring correction between the two cups. The
second term is stated for completeness and is the natural extension where the pattern must fill a substantially wider rim.
 
Scaling the cup also requires scaling the amount of liquid: a larger cup takes a larger espresso dose and more milk. Since flow rate depends on tilt and head height, a jug with more milk pours proportionally more over the same tilt profile, and the demonstrated \emph{timing} transfers unchanged. This is consistent with the observation reported in Section~\ref{sec:results}.

\section{Experimental Setup}
\label{sec:setup}
 
\subsection{Platform and Hardware}
\label{sec:platform}
 
Experiments were performed on a Fairino FR5 six-axis collaborative robot
running ROS~2 Humble on Ubuntu~22.04. Motion planning used MoveIt~2;
inverse kinematics was solved per waypoint for the $\mathcal{E}$ frame defined in Section~\ref{sec:frames},
and trajectories were executed through the \texttt{FollowJointTrajectory}
action interface.
 
The milk jug is attached to the flange by a printed adapter that holds
the jug handle planar with the flange, such that the jug is upright when
the final wrist joint is at zero. Two HTC Vive trackers are used: one
mounted on the jug on the same 1/4" thread used during
demonstration, and one fixed to the demonstration jig as the anchor.
 
The cup sits in a passive swivel whose rotation is measured by an AS5600
magnetic encoder streaming at \qty{100}{\hertz}. For reproduction, the
swivel is driven by a NEMA~17 stepper through a PD~Stepper (ESP32-S3,
TMC2209) controller, which receives the cup-angle profile over WiFi before
each run and executes it from a local timer on trigger; the same AS5600
provides closed-loop confirmation of the executed angle. Phase transitions
are marked during demonstration using a USB foot pedal.
 
\subsection{Latte Preparation}
\label{sec:recipe}
 
Commercial-grade equipment was used for consistent beverage preparation: a
La~Marzocco Linea~Mini espresso machine and an Anfim SCODY grinder. Two
industry-standard cup sizes were used, with the recipe scaled to each so
that the milk-to-espresso ratio and fill level were comparable
(Table~\ref{tab:recipe}). Pattern formation depends on the contrast
between milk and crema and on the depth of liquid in the cup, so the
recipe must scale with the cup.
 
\begin{table}[h]
\centering
\small
\caption{Beverage parameters for the two cup sizes.}
\label{tab:recipe}
\begin{tabular}{lrr}
\toprule
 & \SI{280}{\milli\litre} & \SI{400}{\milli\litre} \\
\midrule
Espresso dose & \SI{21}{\gram} (single) & \SI{42}{\gram} (double) \\
Milk volume   & \SI{250}{\milli\litre}  & \SI{350}{\milli\litre}  \\
Rim diameter  & \SI{80}{\milli\metre}   & \SI{90}{\milli\metre}   \\
\bottomrule
\end{tabular}
\end{table}
 
\subsection{End-to-End Motion Measurement}
\label{sec:measurement}
 
The robot's pour is measured with the same tracker mounted on the same jug
at the same position used during demonstration, so the tracker-to-jug
transform is identical in both recordings and cancels in the comparison,
as do the flange-to-jug offset and the cup placement. The comparison
therefore directly addresses whether the robot's jug moved as the human's
jug moved.
 
The robot recording's phase boundaries are taken from the executed
trajectory's known phase spans, with the two time series aligned by
cross-correlating the commanded and measured speed profiles.
 
\subsection{Experimental Conditions}
\label{sec:conditions}
 
Three conditions were evaluated, each reproducing the same demonstration.
 
\paragraph{E1 --- One-shot CDMP:}
The complete demonstration is encoded as a single CDMP with one parameter
set across all phases ($\alpha_s = 7.0$, $K = 800$, $\rho = 7.0$,
$\tau = 1.0$), serving as the baseline.
 
\paragraph{E2 --- Phased CDMP:}
The demonstration is segmented as described in Section~\ref{sec:phases}
and each phase is encoded with its own $(\alpha_s, K, \rho, \tau)$,
listed in Table~\ref{tab:cdmp-params}. Only \texttt{mix\_to\_art} is
temporally scaled ($\tau = 4$), leaving both pouring phases at
demonstration speed (Section~\ref{sec:cdmp}).
 
\begin{table}[h]
\centering
\small
\caption{Per-phase CDMP parameters: temporal scaling~$\tau$,
basis-function density~$\rho$, stiffness~$K$ ($D = 2\sqrt{K}$).}
\label{tab:cdmp-params}
\begin{tabular}{lrrrr}
\toprule
Phase & $\alpha_s$ & $K$ & $\rho$ & $\tau$ \\
\midrule
\texttt{base\_to\_mix} & 7.0 & 500 & 5.0  & 1.0 \\
\texttt{mixing}        & 7.0 & 800 & 10.0 & 1.0 \\
\texttt{mix\_to\_art}  & 7.0 & 700 & 6.0  & 4.0 \\
\texttt{art}           & 7.0 & 800 & 8.0  & 1.0 \\
\texttt{return\_base}  & 7.0 & 500 & 5.0  & 1.0 \\
\bottomrule
\end{tabular}
\end{table}
 
\paragraph{E3 --- Cup-size adaptation:}
The phased representation of E2 is transferred to the
\SI{400}{\milli\litre} cup by spatial offsetting alone
(Section~\ref{sec:generalisation}), with no re-demonstration and no change
to any parameter in Table~\ref{tab:cdmp-params}. The latte recipe is
scaled as in Table~\ref{tab:recipe}.
 
Conditions E2 and E3 were executed with milk and espresso. E1 was
evaluated on trajectory evidence alone, as the reproduced trajectory
placed the jug outside the cup during the art phase and a liquid trial
was unsafe.
 
\subsection{Phase Evaluation Metrics}
\label{sec:metrics}
 
Metrics are selected per phase type, as success criteria differ across
phases.
 
\paragraph{Transition phases:}
We report the endpoint error, computed over settled windows at each end of
the phase to reduce sensitivity to the exact phase boundary placement.
 
\paragraph{Mixing phase:}
Canvas setting is evaluated by coverage: the mean and maximum radial
extent of the pour footprint about its centroid, and the coverage ratio
between reproduction and demonstration.
 
\paragraph{Art phase:}
We align the reproduced trajectory to the demonstration by translation and
rotation (without scaling) and report the residual Shape RMS and the symmetric
Hausdorff distance. Alignment with uniform scaling is reported separately,
and the pull-through stroke is characterised by its length and heading.
 
\paragraph{Jug tilt:}
Tilt---the angle between the jug's vertical axis and gravity---is reported
for every phase. It is the primary determinant of flow rate and is
directly comparable between recordings without frame calibration
(Section~\ref{sec:measurement}). We report peak tilt, RMS difference, and
the amplitude ratio between reproduction and demonstration.
 
\paragraph{Qualitative assessment:}
The produced pattern is photographed and compared with the demonstration's
pattern.

\section{Results}
\label{sec:results}
 
Results are reported for the three conditions of Section~\ref{sec:conditions}. E1 and E2 are compared quantitatively against the human demonstration. E3 is a CDMP-scaled version of E2 and is assessed visually, for the reason given in Section~\ref{sec:res-cup}.
 
\subsection{Reproduction Fidelity: One-Shot versus Phased}
\label{sec:res-fidelity}
 
\begin{figure}[h]
\centering
\begin{minipage}[t]{0.49\columnwidth}
  \centering
  \includegraphics[width=\linewidth]{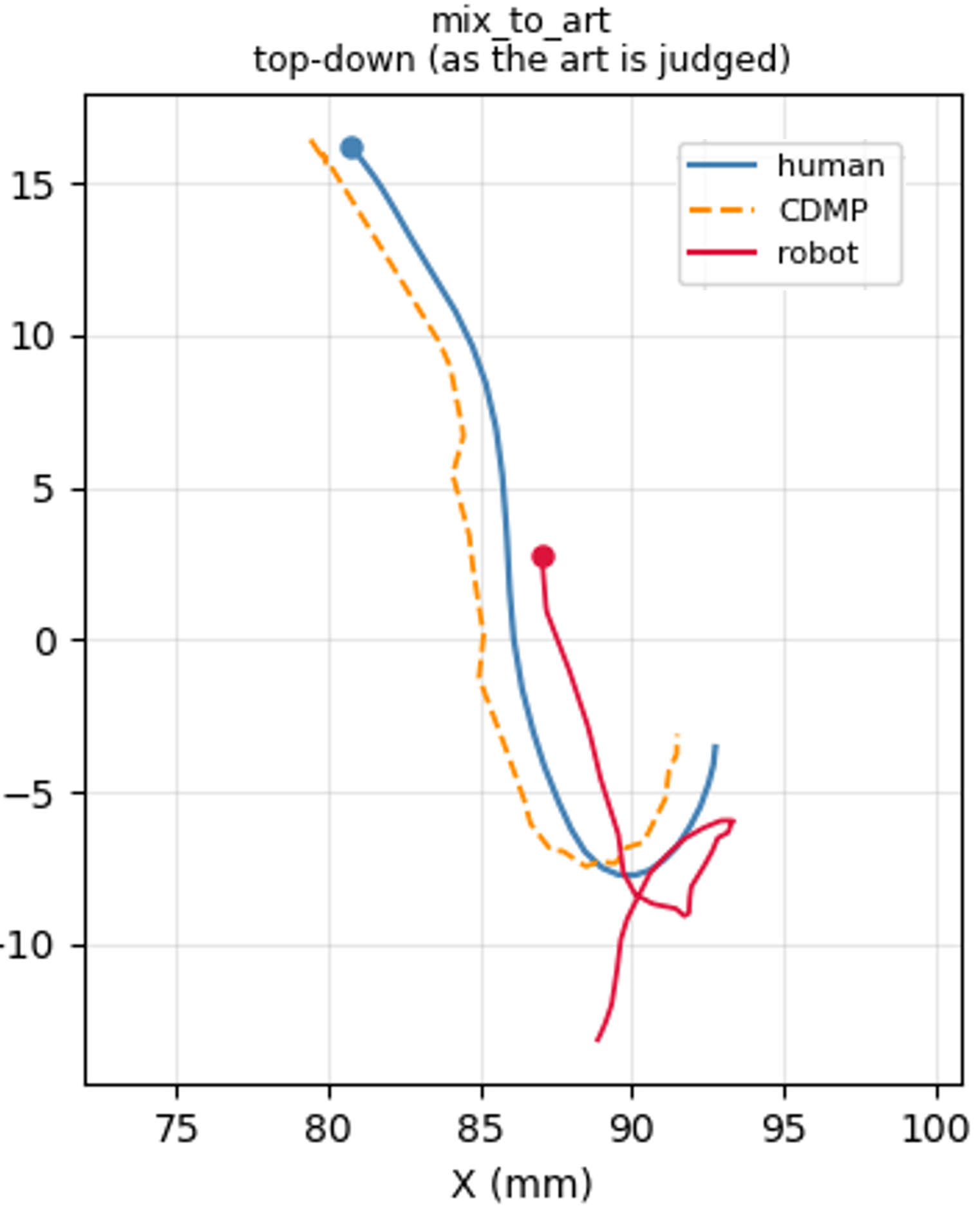}\\
  {\footnotesize (a) E1, one-shot}
\end{minipage}\hfill
\begin{minipage}[t]{0.49\columnwidth}
  \centering
  \includegraphics[width=\linewidth]{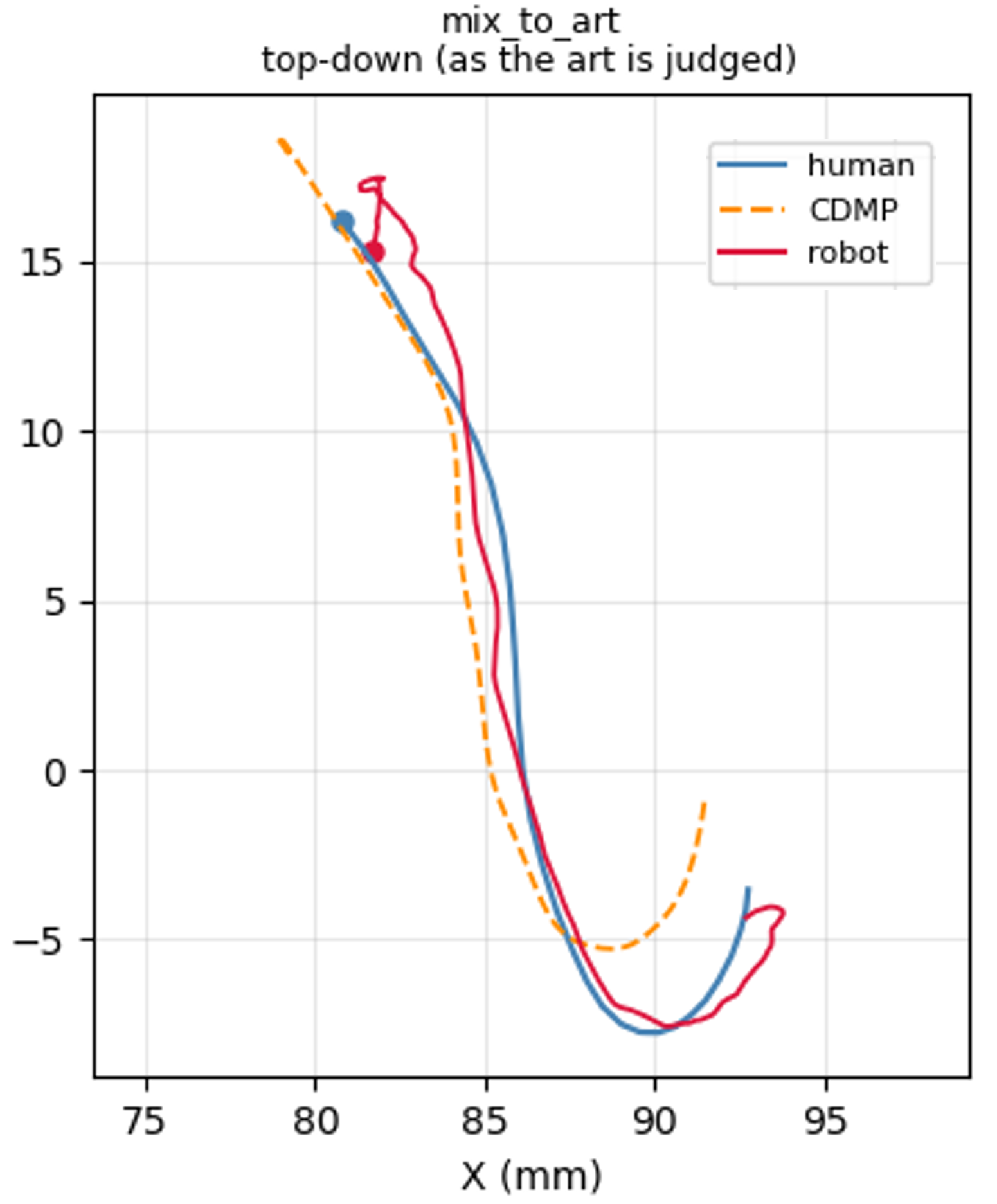}\\
  {\footnotesize (b) E2, phased}
\end{minipage}
\caption{Top-down trajectories of the \texttt{mix\_to\_art} transition for
(a) the one-shot and (b) the phased reproduction, showing the human
demonstration, the CDMP rollout and the executed robot motion.}
\label{fig:topdown}
\end{figure}
 
\begin{table*}[h]
\centering
\caption{Per-phase comparison of one-shot (E1) and phased (E2)
reproduction against the human demonstration. Phases are grouped by type.
$\times$ denotes the improvement factor E1/E2; lower is better except for
coverage ratio, where $1.00$ is ideal.}
\label{tab:quant}
\begin{tabular}{ll l rrrr rrrr}
\toprule
& & & \multicolumn{4}{c}{Phase Evaluation Metric}
  & \multicolumn{2}{c}{Tilt RMS (\si{\degree})}
  & \multicolumn{2}{c}{Height RMS (\si{\milli\metre})} \\
\cmidrule(lr){4-7}\cmidrule(lr){8-9}\cmidrule(lr){10-11}
Phase & Type & Metric & E1 & E2 & $\times$ & Unit
      & E1 & E2 & E1 & E2 \\
\midrule
\texttt{base\_to\_mix} & transition & endpoint error
  & 30.1 & \textbf{9.5}  & 3.2 & mm  & 5.05 & \textbf{1.03} & 22.1 & \textbf{6.3}  \\
\texttt{mix\_to\_art}  & transition & endpoint error
  & 14.9 & \textbf{3.3}  & 4.5 & mm  & 9.51 & \textbf{0.61} & 60.8 & \textbf{18.7} \\
\texttt{return\_base}  & transition & endpoint error
  & 38.3 & \textbf{11.9} & 3.2 & mm  & 9.35 & \textbf{3.54} & 40.7 & \textbf{12.4} \\
\midrule
\texttt{mixing}        & pouring    & coverage ratio
  & 1.18 & 1.17          & 1.0 & ---  & 4.35 & \textbf{0.88} & 18.0 & \textbf{15.9} \\
\texttt{art}           & pouring    & shape RMS
  & 6.35 & \textbf{2.38} & 2.7 & mm  & 3.15 & \textbf{0.75} & 11.3 & 16.5 \\
\bottomrule
\end{tabular}

\vspace{2pt}
\end{table*}
 
Phased reproduction (E2) reduced the primary metric in every phase except mixing (Table~\ref{tab:quant}). Transition endpoint errors fell from 30.1, 14.9 and \SI{38.3}{\milli\metre} to 9.5, 3.3 and
\SI{11.9}{\milli\metre}, a mean reduction from 27.8 to \SI{8.2}{\milli\metre} ($3.4\times$), and the art-stroke shape residual fell from 6.35 to \SI{2.38}{\milli\metre} ($2.7\times$).
 
The largest gain was at \texttt{mix\_to\_art}, the phase given $\tau = 4$ in E2 (Fig.~\ref{fig:topdown}). In E1 this transition carried the largest height and tilt errors of any phase (\SI{60.8}{\milli\metre} and
\SI{9.51}{\degree}); in E2 these fell to \SI{18.7}{\milli\metre} and \SI{0.61}{\degree}. The height component is not visible in the top-down view and is therefore reported in Table~\ref{tab:quant}. Because the art phase begins where this transition ends, errors here determine the starting pose of the pour that forms the pattern.
 
The effect carries into the art stroke itself. E1 produced a stroke $30\%$ longer than demonstrated (size ratio $1.30$, stroke length ratio $1.18$); E2 was within $5\%$ (size ratio $1.05$, length ratio $1.03$). The worst-case local deviation (Hausdorff distance) fell from 10.9 to \SI{6.9}{\milli\metre}.
 
Not every measure improved. Mixing coverage was unchanged between conditions ($1.18$ and $1.17$). In absolute terms this corresponds to a mean footprint radius of \SI{8.6}{\milli\metre} against
\SI{7.4}{\milli\metre} in the demonstration (maximum 18.2 against \SI{15.2}{\milli\metre}), a difference of 1--\SI{3}{\milli\metre} that is small relative to the cup rim, so the canvas is set over essentially the demonstrated area.
\subsection{Jug Tilt and Flow}
\label{sec:res-tilt}
 
\begin{figure}[h]
\centering
\begin{minipage}[t]{0.49\columnwidth}
  \centering
  \includegraphics[width=\linewidth]{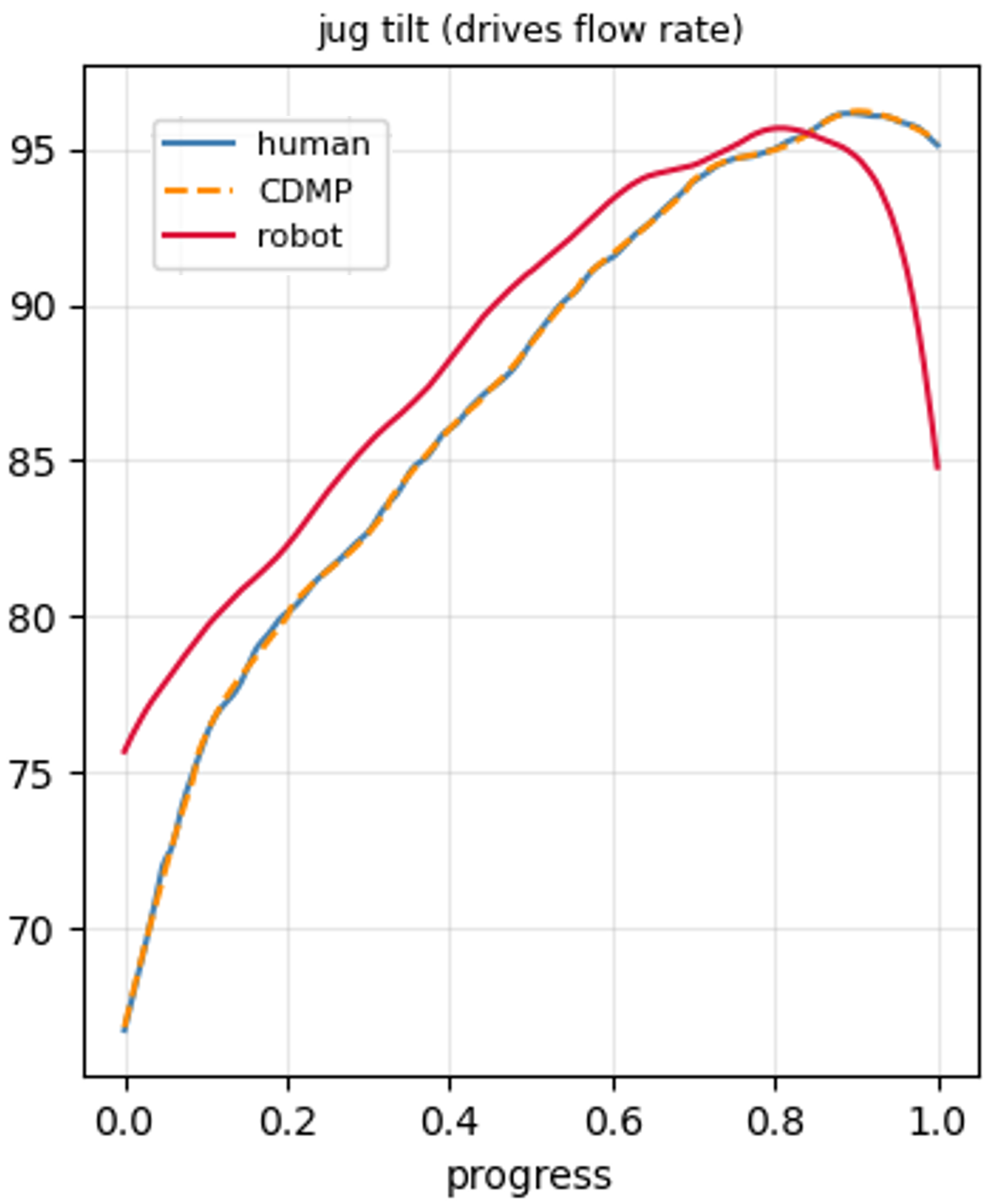}\\
  {\footnotesize (a) E1, one-shot}
\end{minipage}\hfill
\begin{minipage}[t]{0.49\columnwidth}
  \centering
  \includegraphics[width=\linewidth]{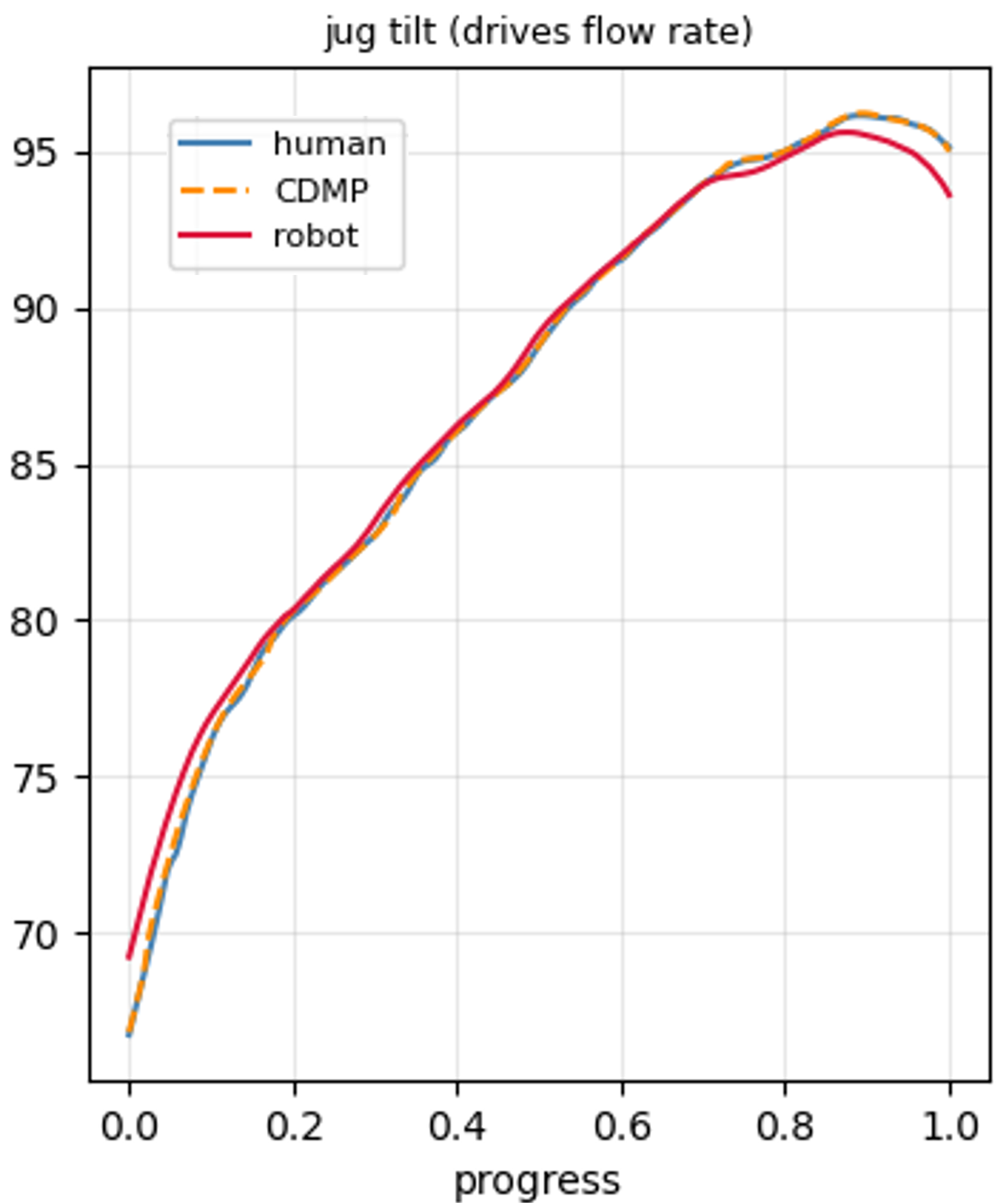}\\
  {\footnotesize (b) E2, phased}
\end{minipage}
\caption{Jug tilt relative to ground plane over the art phase for (a) the
one-shot and (b) the phased reproduction.}
\label{fig:tilt}
\end{figure}
 
Tilt is the primary determinant of flow rate which has a significant impact on the art pattern in addition to position.
 
Over the art phase both conditions reproduced the demonstrated peak tilt (\SI{96.1}{\degree} demonstrated; 95.7 and \SI{95.6}{\degree} for E1 and E2, amplitude ratios $1.00$ and $0.99$), but E2 followed the demonstrated
profile much more closely: tilt RMS fell from 3.15 to \SI{0.75}{\degree} (Fig.~\ref{fig:tilt}). Matching the peak alone does not match the pour, since the poured volume depends on how tilt scales over time.
 
Phasing reduced tilt RMS in every phase, by factors from $2.6$ to $15.6$ (Table~\ref{tab:quant}). The largest change was again at \texttt{mix\_to\_art}, where E1 overshot the demonstrated peak tilt by $29\%$ (\SI{59.3}{\degree} against \SI{46.0}{\degree}) and E2 matched it ($1.01\times$). In the pouring phases of \texttt{mixing} and \texttt{art}, phasing improved by a factor of  $4.9$ and $4.2$  respectively, significantly closing the error between demonstrated and reproduced pour volumes.
\subsection{Qualitative Outcome}
\label{sec:res-visual}
 
\begin{figure}[h]
\centering
\begin{minipage}[t]{0.32\columnwidth}
  \centering
  \includegraphics[width=\linewidth]{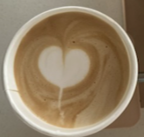}\\
  {\footnotesize (a) Demonstration}
\end{minipage}\hfill
\begin{minipage}[t]{0.32\columnwidth}
  \centering
  \includegraphics[width=1.01\linewidth]{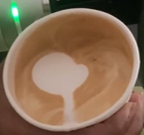}\\
  {\footnotesize (b) E2, \SI{280}{\milli\litre}}
\end{minipage}\hfill
\begin{minipage}[t]{0.32\columnwidth}
  \centering
  \includegraphics[width=0.9\linewidth]{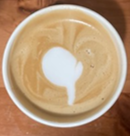}\\
  {\footnotesize (c) E3, \SI{400}{\milli\litre}}
\end{minipage}
\caption{Top-down photographs of the produced latte art. E1 is omitted as
it was not poured with milk (Section~\ref{sec:res-visual}).}
\label{fig:art}
\end{figure}
 
E1 was unsafe to run with milk, given the transition errors
described in Section~\ref{sec:res-fidelity}, and is therefore evaluated on
trajectory evidence alone.
 
The phased reproduction (E2) produced a recognisable heart
(Fig.~\ref{fig:art}b). Compared with the demonstration
(Fig.~\ref{fig:art}a) it was blobbier, with a pull-through that was
displaced from the centreline. The off-centre pull-through is consistent
with the art-phase deviation reported above: the $9.4^{\circ}$ heading
error and the roughly doubled lateral extent of the stroke.
 
\subsection{Generalisation to a Larger Cup}
\label{sec:res-cup}
 
Transferring the phased representation to the \SI{400}{\milli\litre} cup
required a spatial offset of the \texttt{mix\_to\_art} goal and a scaled
recipe (Table~\ref{tab:recipe}). No temporal parameter was altered and no
further demonstration was recorded. The resulting heart was centred in
the larger cup and visually near-identical in character to the
\SI{280}{\milli\litre} result (Fig.~\ref{fig:art}c).
 
The offset acts on the goal
of the transition and the art phase is chained rigidly to it, so the art
stroke's shape is unchanged by construction and only its location moves. The pattern lands centred in
the new cup and provides ground for generalisation to different cup sizes. 
The practical implication is that adapting
to a new cup reduces to a geometric offset and a recipe change rather than
a new demonstration, which is the relevant property for deployment across
the range of cup sizes used in a commercial setting.

\section{Discussion}
\label{sec:discussion}

Decomposing the demonstration into phases improved reproduction in every
phase except mixing. A single CDMP (E1) left transition endpoint errors of
15--\SI{38}{\milli\metre} and an art stroke $1.30\times$ the demonstrated
size. Per-phase CDMPs (E2) reduced transition endpoint error by
$3.2$--$4.5\times$, brought the stroke to $1.05\times$ the demonstrated
size, and reproduced tilt to below $1^{\circ}$ RMS in both pouring phases.
The \texttt{mix\_to\_art} transition lasts \SI{0.70}{\second}, under $4\%$
of the demonstration, so a single CDMP assigns it few basis functions
(Section~\ref{sec:cdmp}), and its error carries into the art phase.

For both E1 and E2, the art phase received nearly the same parameters in both
conditions ($K = 800$ and $\tau = 1$ in each; $\rho = 8$ against $7$), and
its shape residual still fell $2.7\times$. This suggests that
decomposition accounts for much of the improvement.

E3 uses the free start and goal of the CDMP: displacing the goal of
\texttt{mix\_to\_art} relocates the art phase whilst
preserving the art phase. A single offset centred the heart in the larger
cup without re-demonstration. We therefore present phase decomposition as
the practical contribution and goal parameterisation as the mechanism that
makes cup adaptation possible without re-demonstration.

The reproduced heart was blobbier than the demonstration's, with an
off-centre pull-through, despite close agreement in stroke size
($1.05\times$) and tilt profile (RMS \SI{0.75}{\degree}). The
blobbier texture may be attributed to milk texturing as the reproduction milk is visibly foamed heavier. This attribution rests on the match in stroke size and tilt
profile without a controlled comparison. The method is open-loop with
respect to the fluid, so it cannot correct a pour when milk texture
differs.

Two deviations persisted in both conditions. The heading error of the art
stroke ($9.4^{\circ}$ in E2) displaces the pull-through by about
\SI{11}{\milli\metre}, consistent with its off-centre position; the goal
offset of the art phase could correct it. The
mixing footprint is $17\%$ wider than demonstrated (a
\SI{1.2}{\milli\metre} difference in mean radius). A discrete primitive has
no notion of repetition, which may contribute to this.

\section{Limitations and Future Work}
\label{sec:limitations}

The system is open-loop with respect to milk, as texture is not sensed or controlled, and the reproduced heart differed from the demonstration despite closely matching stroke size and tilt profile. Investigating the root cause of this visual discrepancy is a key area of interest, as coffee preparation is multi-factorial and variations in a combination of factors can lead to differences.  

Future work will add three additional features. Vision-based detection of cup position and diameter would allow the spatial offset to be computed from Equation~\eqref{eq:cup-offset} rather than tuned, and would allow the radial scaling term to scale the heart size to the rim. 

A rhythmic DMP for the \texttt{mixing} phase would represent its periodic motion directly rather than as a discrete movement, and would allow for investigation of sequencing periodic and discrete DMPs. 

Finally, learning further patterns such as rosettas and tulips would extend the skill sen beyond the heart, and through DMPs sequencing allow for combinations of different pattern phases to produce truly customisable latte art.

\section{Conclusion}
\label{sec:conclusion}

A one-shot latte art reproduction and adaptation has been presented, in which a barista demonstration
is decomposed into five phases, each encoded as an independent CDMP.
Per-phase encoding reduced transition endpoint error by
$3.2$--$4.5\times$ and art-stroke shape error by $2.7\times$ relative to
a single-CDMP baseline, and reproduced jug tilt to below $1^{\circ}$ RMS
in the pouring phases, yielding a recognisable heart. A spatial
offset transferred the learnt pour to a \SI{400}{\milli\litre} cup
without re-demonstration or temporal retuning.

The results show that the process in which a demonstration is decomposed matters as much
as the choice of primitive: the transition between mixing and art, though
under 4\% of the demonstration duration, determined the starting pose of
the art stroke and was the dominant source of error in the single-CDMP
baseline. Representing the mixing phase, whose motion is periodic, with a
discrete primitive was the one aspect that per-phase encoding did not
improve.

Future work should incorporate sensing to automatically detect cup
geometry and position, enabling the spatial offset to be computed rather
than tuned and extending the skillset to other patterns such as rosettas, tulips, and combination art patterns.

\bibliography{references}
\bibliographystyle{named}

\end{document}